\documentclass[runningheads]{llncs}
\usepackage{graphicx}
\usepackage{multirow}
\usepackage{mathrsfs}
\usepackage{subfig}
\usepackage{tikz}
\usepackage{amsmath,amssymb} 
\usepackage{color}

\usepackage{hyperref}

\begin{document}
\title{DRPN: Making CNN Dynamically Handle\\
Scale Variation}
\author{Jingchao~Peng \and Haitao~Zhao \and Zhengwei~Hu \and Kaijie~Zhao \and Zhongze~Wang}
\institute{East China University of Science and Technology,\\
Automation Department, School of Information Science and Engineering}

\maketitle

\begin{abstract}
Based on our observations of infrared targets, serious scale variation
along within sequence frames has high-frequently occurred. In this
paper, we propose a dynamic re-parameterization network (DRPN) to
deal with the scale variation and balance the detection precision
between small targets and large targets in infrared datasets. DRPN
adopts the multiple branches with different sizes of convolution kernels
and the dynamic convolution strategy. Multiple branches with different
sizes of convolution kernels have different sizes of receptive fields.
Dynamic convolution strategy makes DRPN adaptively weight multiple
branches. DRPN can dynamically adjust the receptive field according
to the scale variation of the target. Besides, in order to maintain
effective inference in the test phase, the multi-branch structure
is further converted to a single-branch structure via the re-parameterization
technique after training. Extensive experiments on FLIR, KAIST, and
InfraPlane datasets demonstrate the effectiveness of our proposed
DRPN. The experimental results show that detectors using the proposed
DRPN as the basic structure rather than SKNet or TridentNet obtained
the best performances.
\end{abstract}

\section{Introduction}

\setlength{\abovecaptionskip}{1mm}
\begin{figure}[t]
\begin{centering}
\textsf{\includegraphics[width=0.9\columnwidth]{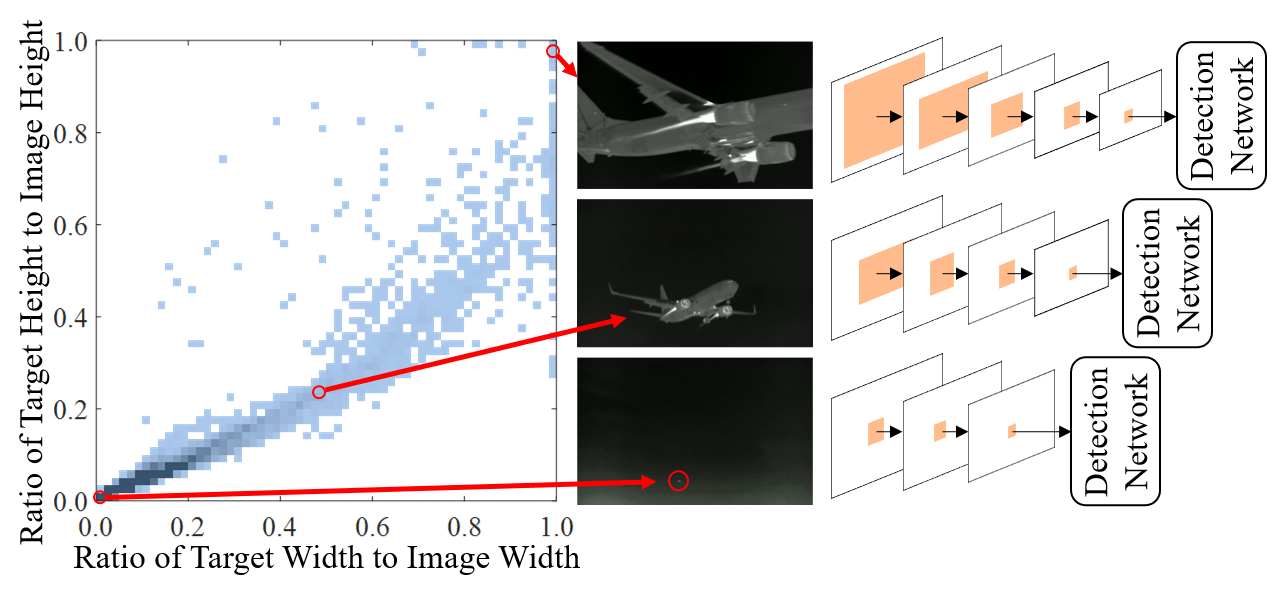}}
\par\end{centering}
\caption{\label{fig:1}The statistical observation of scale variation in infrared
airplane datasets. The area proportion of the small target is less
than 0.1\%; The large target is almost full of the whole image. When
detecting large targets, the receptive field needs to be large, while
detecting small targets, the receptive field needs to be small.}
\vspace{-0.5cm}\end{figure}

Visual target detection is an essential task for many intelligent systems.
Over the past decade, with the development of deep learning, target detection has achieved great success.
Among them, the detection at degraded
visibility conditions or at a great distance is a challenging problem
\cite{smallsurvey1,smallsurvey2,xu1,dense1,rgbt1}. On the contrary, thermal infrared cameras detect the thermal
radiation of targets and convert it into images, which are not influenced
by the changing of the visibility conditions. Furthermore, thermal
infrared cameras have the ultra-long-distance detection capability.
Very distant targets can still be recorded in infrared images. Due
to the good properties of thermal infrared imaging, infrared target
detection methods have been widely studied \cite{map1,map2,xu1,xu2,dense1,dense2,rgbt1,rgbt2,rgbt3}.

The existing infrared target detection methods usually focus on the
detection of small targets, which mostly rely on the fine-resolution
prediction map \cite{map1,map2}. For instance, drop the low-resolution
layers and enhance the high-resolution layer to make networks more
suitable for the detection task of infrared small targets \cite{xu1,xu2};
use the dense connection to keep the high-resolution feature map in
the previous channels \cite{dense1,dense2}; use multimodal feature maps like
combining with the visible spectrum sensor to increase resolution
\cite{rgbt1,rgbt2,rgbt3}. However, these methods pay less attention to scale
variation which has high-frequently occurred in infrared datasets.

When detecting large-size targets, the feature map of the detection
networks usually needs to contain semantic information to enhance
the detection precision. a large receptive field after deep convolutions
often needs to match the size of the target \cite{sepc}. However, when
detecting small targets, the feature map needs to have detailed information
for positioning, a small receptive field should be utilized to prevent
target information loss \cite{small1,small2}. While the large receptive field
(and high-level semantic understanding) is often in conflict with
the small receptive field (and fine-resolution prediction maps) since
deep networks learn more semantic representations by gradually attenuating
the size of the feature maps \cite{resnet}. Besides, as thermal infrared
images have ultra-long-distance detection capabilities, this contradiction
is further exacerbated. For instance, in the visible light target
detection datasets such as COCO, the size of small targets is usually
less than 32\texttimes 32, the size of large targets is greater than
96\texttimes 96, and the size of medium targets is between them. While
in the infrared datasets, the small target is usually less than 9\texttimes 9,
and the large target is more than 256\texttimes 256. The scale variation
in infrared images is much more serious than that of visible images.
To this end, we conduct statistical analysis about scale variation
on a large infrared target detection dataset.

We count infrared targets according to the ratio of target size to
image size, as shown in Fig. \ref{fig:1}. Where the horizontal and vertical
coordinates in the figure represent the ratio of target width to image
width and the ratio of target height to image height, respectively.
Each blue point represents a target. When the image size is 300\texttimes 300,
the size of the smallest target is 9\texttimes 5, the area proportion
is less than 0.1\%. The size of the biggest target is 296\texttimes 292,
which is almost full of the whole image. It can be seen from the figure
that the scale variation of infrared targets along with sequence frames
is very serious. Large scale variation brings the challenge to infrared
target detection. When detecting large targets, the network needs
to be deep to obtain large receptive fields and high-level semantic
information; when detecting small targets, the network needs to be
shallow, so as not to reduce the resolution of the prediction map
and lose the target information.

Just motivated by the above observations, in this paper, we propose
a dynamic re-parameterization network (DRPN) to deal with the scale
variation and balance the detection precision of small targets and
large targets. DRPN adopts a multi-branch structure, in which every
branch has different receptive fields. Besides, DRPN uses the strategy
of dynamic convolution, and automatically generates weights according
to the input, so that the DRPN can automatically weight different
branches to automatically adjust the receptive field according to
the changing of the target size. In this way, DRPN solves the problem
of network design contradiction caused by scale variation in infrared
target detection. Multi-branch structure brings more memory size and
computation. In order to maintain efficient inference after training,
DRPN adopts re-parameterization technology, which converts the architecture
from multi-branch to single-branch via transforming convolution kernels
of different sizes to convolution kernels of the same size in the
inference phase. Extensive experiments demonstrate that our DRPN is
superior to the baseline as well as other methods.

In summary, our contributions are summarized below:
\begin{enumerate}
\item By combining multi-branch structure and dynamic convolution strategy,
DRPN can automatically adjust the receptive field according to the
scale variation of targets.
\item DRPN adopts multi-branch structure to increase the model capacity
while maintaining effective inference through re-parameterization
technology.
\item The effectiveness of DRPN is demonstrated on FLIR, KAIST, and our
InfraPlane dataset. Detectors using the proposed DRPN as the basic
structure rather than SKNet or TridentNet to replace standard convolution
obtained the best results.
\end{enumerate}

\section{Related Work}
\subsection{Visible and Infrared Target Detection}

Deep-learning-based visible target detection methods can be classified
as anchor-based methods and anchor-free methods. The anchor-based
methods regard the target detection problem as the classification
and regression of candidate regions. The candidate region here is
anchor, which is generated by sliding windows in one-stage methods,
such as SSD \cite{ssd} and YOLO \cite{yolo}; or generated by region proposal
network (RPN) in two-stage methods, such as Faster-RCNN \cite{fasterrcnn},
Cascade-RCNN \cite{cascadercnn}, and so on. In contrast to anchor-based architectures
that rely on anchors for the localization of targets by introducing
a lot of hyperparameters in the model, anchor-free architectures predict
the key points or borders of the target directly, including CornerNet
\cite{cornernet}, FCOS \cite{fcos}, CenterNet \cite{centernet}, to name a few. However,
these methods are designed mainly for medium and large targets, have
no special design for large scale variation in infrared target detection.

Deep-learning-based infrared target detection methods mostly aim at
dim-small target detection. Ding et al. \cite{xu2} improved the network
architecture of SSD for infrared small target detection. Since high-level
semantic layers can hardly extract any feature of dim-small targets,
and these layers make no contribution to the detection of infrared
small targets, Ding et al. first remove high-level semantic layers,
and then enhance the low-level layer by dilated convolution. Dilated
convolution has the advantage of improving the receptive field without
reducing the resolution of the feature maps. Dai et al. \cite{dai1} propose
an asymmetric contextual modulation, with particular emphasis on exchanging
high-level semantics and subtle low-level details. They adopt the
global channel attention modulation to propagate high-level semantic
information down to shallow layers, whereas utilize pixel-wise channel
attention modulation to preserve and highlight infrared small targets
in high-level features. However, both of these two methods are designed
for infrared small target detection and pay less attention to the
scale variation problem.

\subsection{Methods for Handling Scale Variation}

There have also been methods considering the scale variation problem.
The multi-scale image pyramid \cite{pyramid} is a common scheme, but it
will bring a heavy computation burden. Instead of taking multiple
images as input, SSD \cite{ssd} utilizes multi-level features of different
spatial resolutions to alleviate scale variation. FPN \cite{fpn} further
introduces a top-down pathway and lateral connections to enhance the
semantic representation of low-level features at the bottom layers.
SEPC \cite{sepc} captures the inter-scale interactions through an explicit
convolution in the scale dimension, forming a 3-D convolution in the
feature pyramid. These methods study how to make better use of high-level
semantic and fine-resolution features from the way of information
flow. Experiments demonstrate the effectiveness of these methods.

Other methods \cite{sknet,trident} try to change the receptive field of the
neural networks by the idea of multi-branch, which can also deal with
the scale variation problem. SKNet \cite{sknet} uses SoftMax attention
to fuse multiple branches with different kernel sizes. Different attentions
on these branches yield different sizes of the receptive fields. However,
multiple branches represent multiple convolutions, which largely increases
the computational burden. ACNet \cite{acnet} and RepVGG \cite{repvgg} use the
re-parameterization technique to reduce inference time. Specifically,
the training-time model has a multi-branch structure, while the test-time
model converts the architecture from multi-branch to single-branch
via re-parameterization. But they use the same weights when convert
multi-branch to single-branch, which is not suitable to scale variation
problem. TridentNet \cite{trident} generates multiple parallel branches
with different dilation rates, thus endowing the same representational
power for targets of different scales. However, these methods have
no consideration of the dynamic design for the multiple branches.
During the test time, these methods cannot adjust adaptively to handle
the significant scale variation in infrared target detection.
\begin{figure*}[!t]
\begin{centering}
\textsf{\includegraphics[width=1\textwidth]{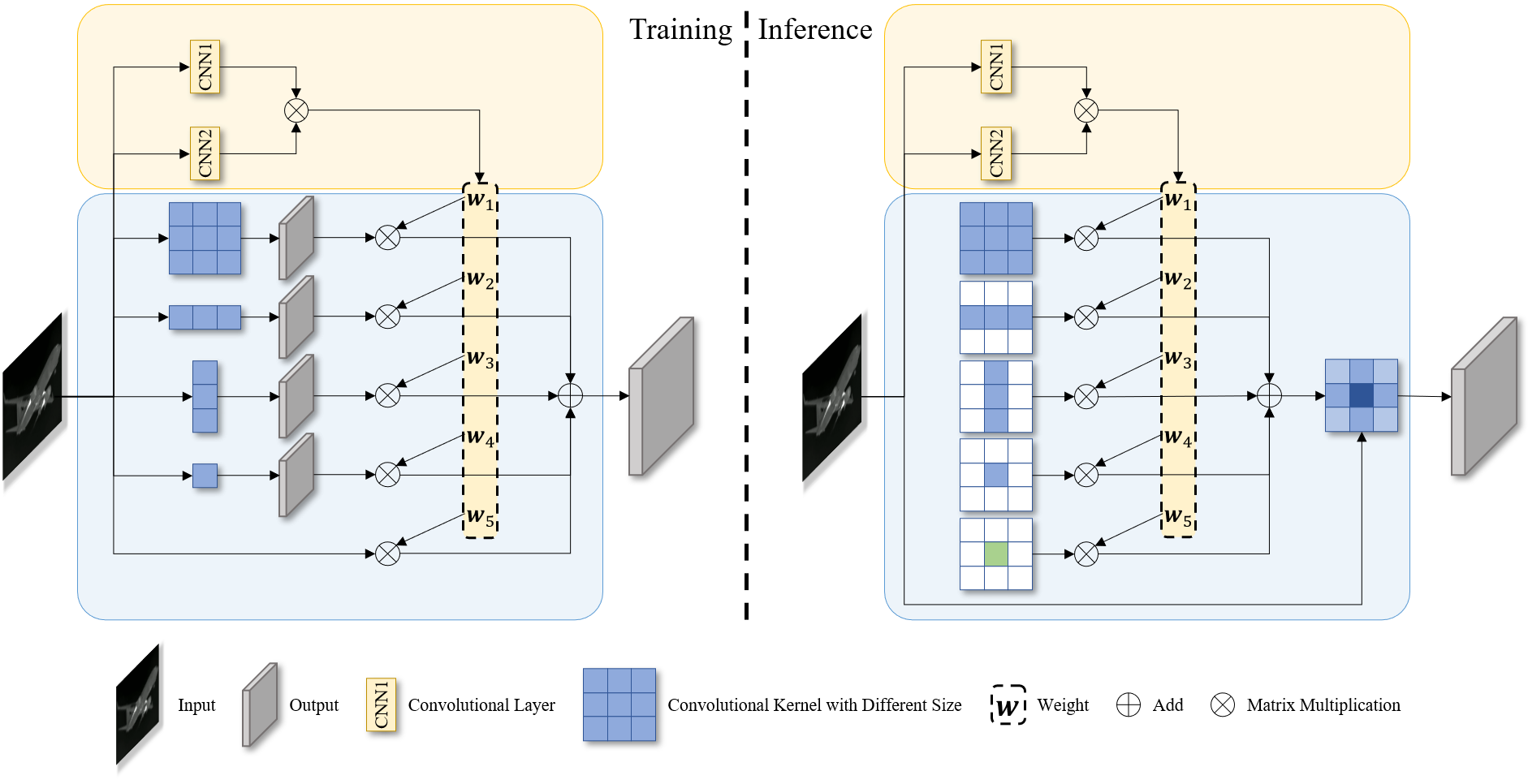}}
\par\end{centering}
\caption{\label{fig:2}Illustration of the proposed DRPN. DRPN consists of
two modules, one is the weight generation module, which is shown in
the yellow box; the other is the multi- branch module, which is shown
in the blue box. The training phase of the multi-branch module is
different from the inference phase. In the training phase, the input
tensor is convoluted first, and then the outputs is weighted and summed;
In the inference phase, the convolution kernel is weighted and summed
first, and then the input tensor is convoluted.}
\vspace{-0.5cm}\end{figure*}

\section{Methodology\label{sec:3}}
In this section, we first overview the architecture of dynamic re-parameterization
network (DRPN) for infrared target detection. Then we will introduce
the two modules of DRPN: the weight generation module and the multi-branch
module.

\subsection{Overall Structure}
The overall network architecture of our proposed DRPN is shown in
Fig. \ref{fig:2}. DRPN aims to solve the scale variation problem in infrared
target detection. DRPN adopts a multi-branch structure, including
four convolution branches whose kernels with different sizes and a
shortcut branch. The convolution kernels with different sizes have
different receptive fields. Meanwhile, the multi-layer structure of
the CNN can further enhance the difference. In order to enable CNNs
to dynamically adjust the receptive field according to the variation
of the target size, DRPN uses the structure of dynamic convolution,
and adaptively generates weights according to the target size of the
input.

DRPN consists of two modules, one is the weight generation module,
the other is the multi-branch module. The weight generation module
generates five weights from the input, corresponding to five branches
of the multi-branch module. During the training phase, each branch
of the multi-branch module convolves the input tensor respectively,
and the obtained five output tensors are weighted and summed according
to the generated weights to obtain the final output tensor. In the
inference phase, the convolution kernels of five branches, including
four convolution branches and a shortcut branch, are unified into
the size of 3\texttimes 3. Then these five convolution kernels weigh
and sum according to generated weights to obtain the final convolution
kernel. Finally, the final convolution kernel is convoluted with the
input tensor to obtain the output tensor.

\begin{figure}[!tbp]
\begin{centering}
\textsf{\includegraphics[width=0.5\columnwidth]{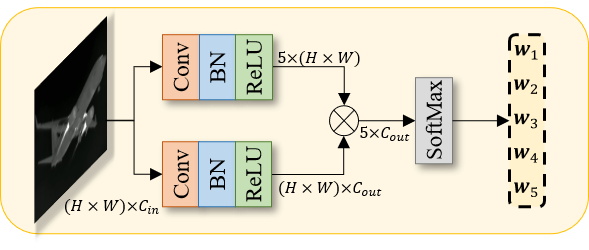}}
\par\end{centering}
\caption{\label{fig:3}Illustration of weight generation module, which is used
to calculate attention matrices suitable for the size of the target
in the infrared image.}
\vspace{-0.5cm}\end{figure}

\subsection{Weight Generation Module}
Given an input tensor $X_{in}\in\mathbb{R}^{H\times W\times C_{in}}$,
and an output tensor $X_{out}\in\mathbb{R}^{H\times W\times C_{out}}$,
the weight generation module first uses two convolution layers to
encode the input tensor into query tensor $Q\in\mathbb{R}^{(H\times W)\times5}$
and key tensor $K\in\mathbb{R}^{(H\times W)\times C_{out}}$:\vspace{-3mm}
\begin{equation}
Q=f_{1}(X_{in})
\end{equation}\vspace{-5mm}
\begin{equation}
K=f_{2}(X_{in}),
\end{equation}
where $f_{1}$ and $f_{2}$ are two different convolutional layers.
To conveniently define the matrix computation, we stack spatial positions
in a row-by-row way: $[X]:\mathbb{R}^{H\times W\times C}\rightarrow\mathbb{R}^{N\times C}$,
where $N=H\times W$ is the number of vertices. Then the attention
matrix $W=[\boldsymbol{w}_{1},\boldsymbol{w}_{2},\boldsymbol{w}_{3},\boldsymbol{w}_{4},\boldsymbol{w}_{5}]^{T}\in\mathbb{R}^{5\times C_{out}}$
is computed by the inner product of the query tensor and the key tensor,
which can be formulated as:
\begin{equation}
W=[\boldsymbol{w}_{1},\boldsymbol{w}_{2},\boldsymbol{w}_{3},\boldsymbol{w}_{4},\boldsymbol{w}_{5}]^{T}=S([Q]^{T}\times[K])
\end{equation}
where $\times$ denote matrix multiplication, and $S(\cdot)$ is the
SoftMax operation. The weight generation module is shown in Fig. \ref{fig:3}.

\begin{table*}[!t]
\caption{\label{tab:1}Classic networks and corresponding special cases of
DRPN.}

\noindent \centering{}%
\begin{tabular*}{1\textwidth}{@{\extracolsep{\fill}}ccc}
\hline
\noalign{\vskip0.01\textwidth}
The Convolutional Layer & \multirow{2}{*}{Weights of DRPN} & \multirow{2}{*}{Structure of DRPN}\tabularnewline
of Classic Network &  & \tabularnewline
\hline
VGG & $\begin{array}{c}
\text{\ensuremath{\boldsymbol{w}_{1}}=\ensuremath{\boldsymbol{1}},}\\
\boldsymbol{w}_{2}=\boldsymbol{w}_{3}=\boldsymbol{w}_{4}=\boldsymbol{w}_{5}=\boldsymbol{0}
\end{array}$ & \centering\raisebox{-.5\height}{\includegraphics[width=0.3\textwidth]{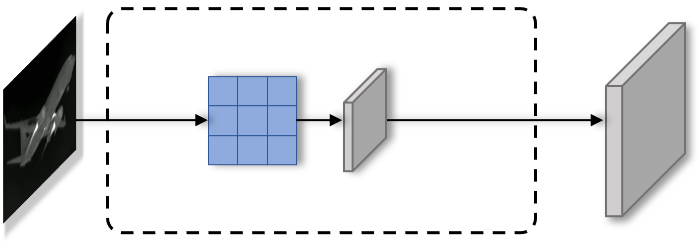}}\tabularnewline
ResNet & $\begin{array}{c}
\boldsymbol{w}_{1}=\boldsymbol{w}_{5}=\frac{1}{2}\times\boldsymbol{1},\\
\boldsymbol{w}_{2}=\boldsymbol{w}_{3}=\boldsymbol{w}_{4}=\boldsymbol{0}
\end{array}$ & \centering\raisebox{-.5\height}{\includegraphics[width=0.3\textwidth]{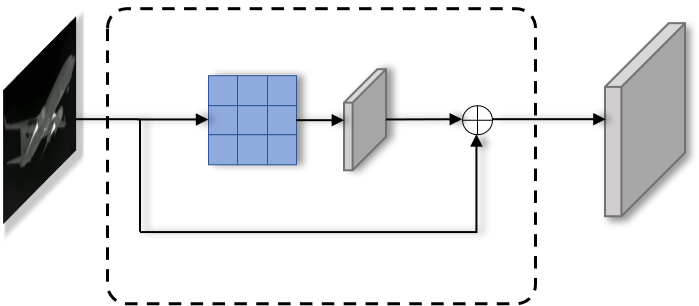}}\tabularnewline
RepVGG & $\begin{array}{c}
\boldsymbol{w}_{1}=\boldsymbol{w}_{4}=\boldsymbol{w}_{5}=\frac{1}{3}\times\boldsymbol{1},\\
\boldsymbol{w}_{2}=\boldsymbol{w}_{3}=\boldsymbol{0}
\end{array}$ & \centering\raisebox{-.5\height}{\includegraphics[width=0.3\textwidth]{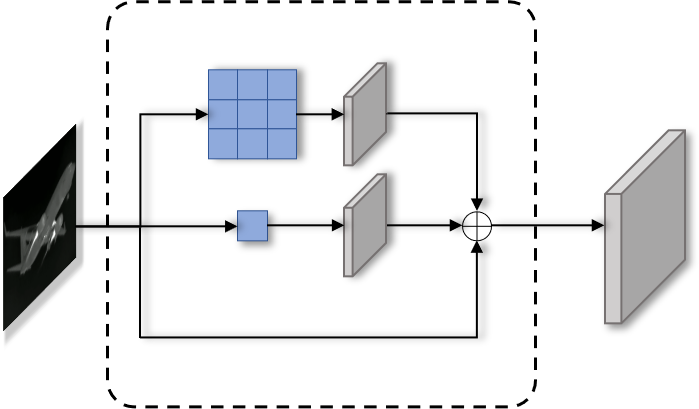}}\tabularnewline
Lightweight Network & $\begin{array}{c}
\boldsymbol{w}_{5}=\boldsymbol{1},\\
\boldsymbol{w}_{1}=\boldsymbol{w}_{2}=\boldsymbol{w}_{3}=\boldsymbol{w}_{4}=\boldsymbol{0}
\end{array}$ & \centering\raisebox{-.5\height}{\includegraphics[width=0.3\textwidth]{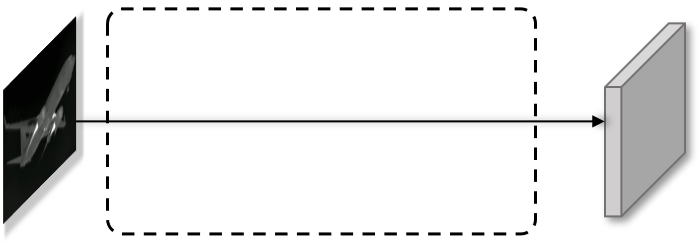}}\tabularnewline
\hline
\end{tabular*}
\vspace{-0.5cm}\end{table*}

\subsection{Multi-branch Module}
Inspired by the re-parameterization technique \cite{acnet,repvgg} and dynamic
convolution \cite{condconv,dyconv,weightnet,dynet}, we construct the multi-branch module by simply
replacing every $3\times3$ layer with five parallel layers with kernel
size $3\times3$, $1\times3$, $3\times1$, $1\times1$, and a shortcut
branch (only if the dimensions of input tensor and output tensor match)
to model the information flow respectively. Since multi-branch architecture
makes one convolution an implicit ensemble of five convolutions, it
is beneficial to training \cite{repvgg34} and has more representation power
\cite{dyconv}. In addition, as will be shown in Sec. \ref{subsec:4.4}, because convolution
kernels of different sizes have different receptive fields, the module
can adjust the receptive field according to the changing of the target
size. For the reason that the high computational density of $3\times3$
convolution \cite{repvgg} and the receptive field of existing networks
is large enough to detect large targets, we do not use $5\times5$
and $7\times7$ convolutions which are used in SKNet \cite{sknet} instead.

At the training phase, first, each branch convolutes the input tensor
respectively, and then the obtained five output tensors are weighted
and summed:\vspace{-3mm}
\begin{equation}
\begin{array}{c}
X_{out}=\boldsymbol{w}_{1}\times_{3}\mathscr{\mathcal{F}}_{3\times3}(X_{in})+\boldsymbol{w}_{2}\times_{3}\mathcal{F}_{1\times3}(X_{in})\\
+\boldsymbol{w}_{3}\times_{3}\mathcal{F}_{3\times1}(X_{in})+\boldsymbol{w}_{4}\times_{3}\mathcal{F}_{1\times1}(X_{in})+\boldsymbol{w}_{5}\times_{3}X_{in}
\end{array},
\vspace{-1mm}\end{equation}
where $\mathscr{\mathcal{F}}_{3\times3}$, $\mathcal{F}_{1\times3}$,
$\mathcal{F}_{3\times1}$, and $\mathcal{F}_{1\times1}$ are the convolution
layers of the corresponding kernel with the size of $3\times3$, $1\times3$,
$3\times1$, and $1\times1$, respectively; $\times_{n}$ refers to
$n$-mode multiplication \cite{dynet2_6}, e.g. $\mathbb{R^{\mathit{i\times j}}\times_{\textrm{3}}R^{\mathit{k\times h\times j}}\rightarrow R^{\mathit{k\times h\times i}}}$. We train the multi-branch module
using the same configurations as the original convolution layers without
tuning any hyper-parameters.

When the training is completed, we convert the multi-branch to a single
convolutional layer through padding zeros outside the $3\times1$,
$1\times3$, and $1\times1$ convolution kernels and making them $3\times3$
kernels, respectively. This transformation also applies to the shortcut
branch because it can be viewed as a $1\times1$ convolution layer
with an identity matrix as the kernel, more details can be referred
in \cite{repvgg}. Then we obtain the final convolution kernel $k_{final}$
by adding up the five convolution kernels:\vspace{-3mm}
\begin{equation}
\begin{array}{l}
k_{final}=\boldsymbol{w}_{1}\times_{1}k_{3\times3}+\boldsymbol{w}_{2}\times_{1}k_{1\times3}+\boldsymbol{w}_{3}\times_{1}k_{3\times1}
+\boldsymbol{w}_{4}\times_{1}k_{1\times1}+\boldsymbol{w}_{5}\times_{1}k_{id}
\end{array},
\end{equation}
where $k_{3\times3}$, $k_{1\times3}$, $k_{3\times1}$, $k_{1\times1}$,
and $k_{id}$ are the kernels of the corresponding filter at the $3\times3$,
$1\times3$, $3\times1$, $1\times1$ convolution layers, and shortcut
layers, respectively. Finally, the output tensor is obtained by convolution
with the final kernel:\vspace{-2mm}
\begin{equation}
\widetilde{X}_{out}=\mathcal{F}_{final}(X_{in}),
\vspace{-2mm}\end{equation}
where $\mathcal{F}_{final}$ is the convolution layer of the final
convolution kernel $k_{final}$. The principle of this conversion
is that the output from convolution with several convolution kernels
which have different sizes is equal to the output obtained from an
equivalent kernel convoluted with the same input, i.e., $X_{out}=\widetilde{X}_{out}$.

With the weight generation module and the multi- branch module, DRPN
can evolve many classical convolutional layers through the combination
of different weights and different branches, as shown in Tab. \ref{tab:1}. Let
$\boldsymbol{1}\in\mathbb{R}^{C_{out}}$ which denote a column vector
where each element is equal to one, and $\boldsymbol{0}\in\mathbb{R}^{C_{out}}$
which denote a column vector where each element is equal to zero.
When $\boldsymbol{w}_{1}=\boldsymbol{1}$, $\boldsymbol{w}_{2}=\boldsymbol{w}_{3}=\boldsymbol{w}_{4}=\boldsymbol{w}_{5}=\boldsymbol{0}$,
DRPN can be regarded as the convolutional layer of VGG \cite{vgg}; when
$\boldsymbol{w}_{1}=\boldsymbol{w}_{5}=1/2\times\boldsymbol{1}$,
$\boldsymbol{w}_{2}=\boldsymbol{w}_{3}=\boldsymbol{w}_{4}=\boldsymbol{0}$,
DRPN can be regarded as the convolutional layer of ResNet \cite{resnet};
when $\boldsymbol{w}_{1}=\boldsymbol{w}_{4}=\boldsymbol{w}_{5}=1/3\times\boldsymbol{1}$,
$\boldsymbol{w}_{2}=\boldsymbol{w}_{3}=\boldsymbol{0}$, DRPN can
be regarded as the convolutional layer of RepVGG \cite{repvgg}; when $\boldsymbol{w}_{5}=\boldsymbol{1}$,
$\boldsymbol{w}_{1}=\boldsymbol{w}_{2}=\boldsymbol{w}_{3}=\boldsymbol{w}_{4}=\boldsymbol{0}$,
DRPN can be regarded as a simplified lightweight convolutional layer
mainly designed to detect small targets. If the weight parameters
$\boldsymbol{w}_{1}$, $\boldsymbol{w}_{2}$, $\boldsymbol{w}_{3}$,
$\boldsymbol{w}_{4}$, and $\boldsymbol{w}_{5}$ are dynamically computed
according to the changing of the target size, the multi-branch module
should be adaptable to the scale variation of the target.
\section{Experiments}

In this section, we first describe the datasets and implementation
details of DRPN and training settings in Sec. \ref{subsec:4.1}. Next,
Sec. \ref{subsec:4.2} compares the results of DRPN with other methods
on three datasets, and Sec. \ref{subsec:4.3-1} show the portability
and generalizability of DRPN. Then, in Sec. \ref{subsec:4.3} We conduct thorough
ablation experiments to demonstrate the effectiveness of the proposed
method. Finally, qualitative performances give the visual comparisons
of our proposed DRPN in Sec. \ref{subsec:4.4}.

\subsection{Datasets and Implementation Details\label{subsec:4.1}}

The methods compared in this section are all pre-trained on the COCO
dataset \cite{coco}. COCO is a large and rich visible image dataset for
object detection and segmentation, which has 330K images, 80 object
categories, and 5 labels per image. FLIR and KAIST \cite{kaist} datasets
are commonly used open-source datasets for infrared target detection.
The FLIR dataset contains 14K images with 10K from short video segments
and random image samples, plus 4K images from a 140-second video.
The KAIST dataset contains a total of 90K images, each containing
both RGB color and infrared versions. It contains a total of 100K
dense annotations. The dataset captured various conventional traffic
scenes including campus, streets, and rural areas during the daytime
and at night. The InfraPlane dataset contains 30,211 infrared images
taken by infrared and hyperspectral cameras, in which airplanes take
off and land near the airport during the day and night. There are
20,177 images in the daytime and 10,034 images in the night. The resolution
of the image is 480\texttimes 300. The examples of the InfraPlane
dataset are shown in Sec. \ref{subsec:4.4}.

We implement our proposed DRPN on the PyTorch platform with I5-10700K @5.0GHz
CPU and NVIDIA TITAN RTX GPU. We adopt SSD \cite{ssd} as our baseline
method in mmdetection \cite{mmdetection}. We replace the corresponding convolution
in conv1\_1 \textasciitilde{} conv5\_3 with our proposed DRPN. The
input images are resized to 300\texttimes 300. Random crop and random
flip are adopted during training. By default, the batch size is 8,
and 24 epochs are trained with the learning rate starting from 0.01
and decreasing by a factor of 0.1 after the 16th and 22th epoch. For
the evaluation, we report the standard COCO evaluation metric of Average
Precision (AP) as well as AP{\scriptsize{}50}, AP{\scriptsize{}s},
AP{\scriptsize{}m}, and AP{\scriptsize{}l}.

\begin{table}[!t]
\caption{\label{tab:2}Comparisons of different target detection methods which
were evaluated on FLIR and KAIST datasets.}

\centering{}%
\begin{tabular}{cc|cc|ccc}
\hline
Dataset  & Method  & AP  & AP50  & APs  & APm  & APl \tabularnewline
\hline
\multirow{4}{*}{FLIR} & SSD  & 21.1  & 49.8  & 11.0  & 27.9  & 36.4 \tabularnewline
 & SKNet  & 22.3  & 51.8  & 10.9  & 29.2  & 39.8 \tabularnewline
 & TridentNet  & 21.6  & 51.2  & 10.7  & 27.2  & 40.1 \tabularnewline
 & DRPN(Ours)  & 22.9  & 52.2  & 11.4  & 29.1  & 38.2 \tabularnewline
\hline
\multirow{4}{*}{KAIST} & SSD  & 29.8  & 63.5  & -  & 29.5  & 42.3 \tabularnewline
 & SKNet  & 25.8  & 57.8  & -  & 25.4  & 35.9 \tabularnewline
 & TridentNet  & 29.7  & 60.6  & -  & 28.7  & 43.1 \tabularnewline
 & DRPN(Ours)  & 32.0  & 63.4  & -  & 31.0  & 45.8 \tabularnewline
\hline
\multirow{4}{*}{InfraPlane} & SSD  & 37.4  & 79.6  & 33.6  & 63.5  & 81.2 \tabularnewline
 & SKNet  & 39.1  & 81.6  & 35.0  & 66.1  & 83.5 \tabularnewline
 & TridentNet  & 37.1  & 80.1  & 33.3  & 62.0  & 85.0 \tabularnewline
 & DRPN(Ours)  & 40.0  & 82.7  & 36.2  & 64.3  & 85.2 \tabularnewline
\hline
\end{tabular}
\vspace{-0.5cm}\end{table}
\subsection{Comparison with Other Methods\label{subsec:4.2}}
We evaluate DRPN on three datasets and compare DRPN with baseline
(SSD \cite{ssd}), SKNet \cite{sknet}, and TridentNet \cite{trident}. DRPN improves
significantly over other methods. Here we report the results in Tab.
\ref{tab:2}. It can be found that DRPN achieves the best performance
in each dataset. Specifically, DRPN achieves 22.9, 32.0, and 40.0
AP, which is 1.8, 2.2, and 2.6 higher than the baseline (SSD). To
compare with SKNet and TridentNet , we adopt the same training settings
and network architecture with DRPN, just the corresponding DRPN is
replaced by SKNet or TridentNet. DRPN improves significantly over
other methods, which improves 1.8, 2.2, and 0.9 AP than the second-best
method on three datasets, respectively. It shows the effectiveness
of dynamic network architecture generated by DRPN with the same set
of parameters.

Furthermore, DRPN achieves the performance with 59.55 GFLOPs computation.
By contrast, the computations of SSD, SKNet, and TridentNet are 34.42,
263.41, and 34.42 GFLOPs, respectively. The computation of DRPN is
25.13 GFLOPs higher than that of baseline (SSD). This computation
cost comes mainly from the weight generation module, which calculates
the weights of different convolution kernels according to the size
of the target in the input images. Nevertheless, given the performance
of DRPN, this computation cost is acceptable.

Compared with the KAIST dataset, the FLIR and the InfraPlane dataset
have the problem of serious scale variation. SKNet adopts the multi-branch
structure both in the training phase and the test phase. Although
the SKNet is more complex, the performance is better than TridentNet.
On the contrary, the target size in the KAIST dataset is relatively
large compared with the other two datasets, and there is no small
target. The dilation rate and the receptive field of TridentNet are
larger than SKNet, therefore TridentNet is more suitable for the KAIST
dataset and has a better performance. While our proposed DRPN outperforms
than SKNet and TridentNet on three datasets, indicating the effectiveness
of the proposed DRPN.

\subsection{Portability and Generalizability \label{subsec:4.3-1}}
\begin{figure}[!t]
\begin{centering}
\subfloat[Experiments on COCO Dataset]{\begin{centering}
\includegraphics[width=0.4\textwidth]{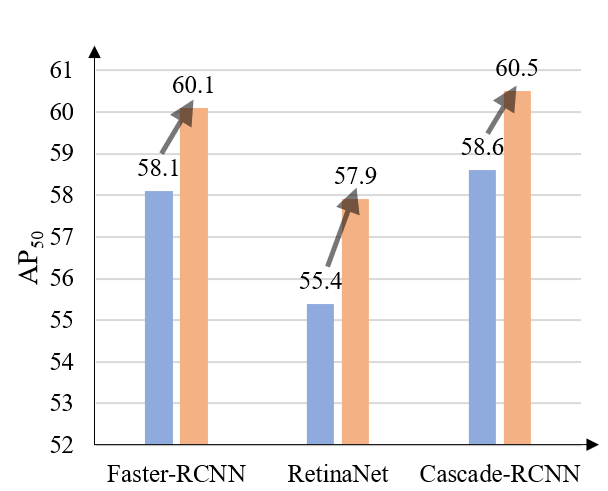}
\par\end{centering}
}\subfloat[Experiments on InfraPlane Dataset]{\centering{}\includegraphics[width=0.479\textwidth]{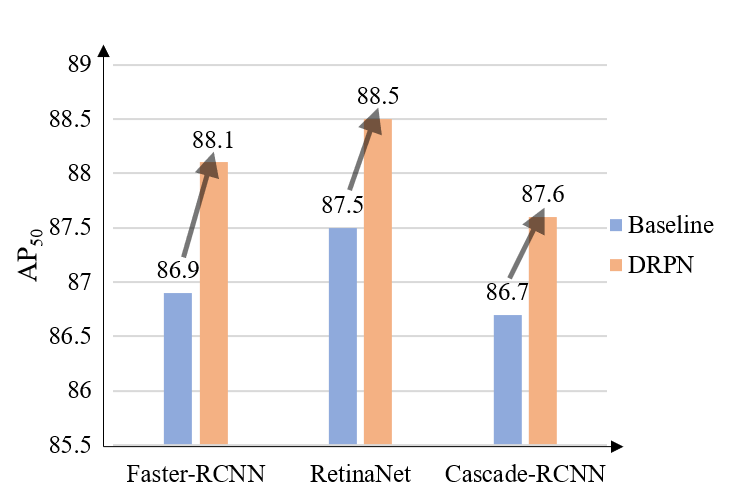}}
\par\end{centering}
\caption{\label{fig:6-1}Experiments on COCO and InfraPlane Datasets using
other detection frameworks.}
\vspace{-0.5cm}\end{figure}
DRPN has a good portability and good Generalizability. That says DRPN
can easily be implemented on other detection frameworks and adapted
to other datasets. Here we implement DRPN on three different detection
frameworks including Faster-RCNN, RetinaNet, and Cascade-RCNN as an
illustration to demonstrate the good portability, and use two datasets
containing COCO and InfraPlane datasets to show the generalization
ability of DRPN to visible and infrared images. We all use ResNet-50
as the backbone. The results are shown in Fig. \ref{fig:6-1}, which
can be shown from the figure that in different frameworks and datasets,
the performances of DRPN are better than those of the baselines (origin
Faster-RCNN, origin RetinaNet, or origin Cascade-RCNN). In specific,
when using Faster-RCNN as the detection framework, DRPN acquires 1.9
and 1.2 AP higher than baseline (origin Faster-RCNN) in COCO
and InfraPlane datasets, respectively. When using RetinaNet as the
detection framework, the increased detection rates are 2.5 and
1.0 AP, and when using Cascade-RCNN as the detection framework, the
two numbers are 1.9 and 0.9 AP.

\begin{figure*}[!t]
\begin{centering}
\textsf{\includegraphics[width=0.9\textwidth]{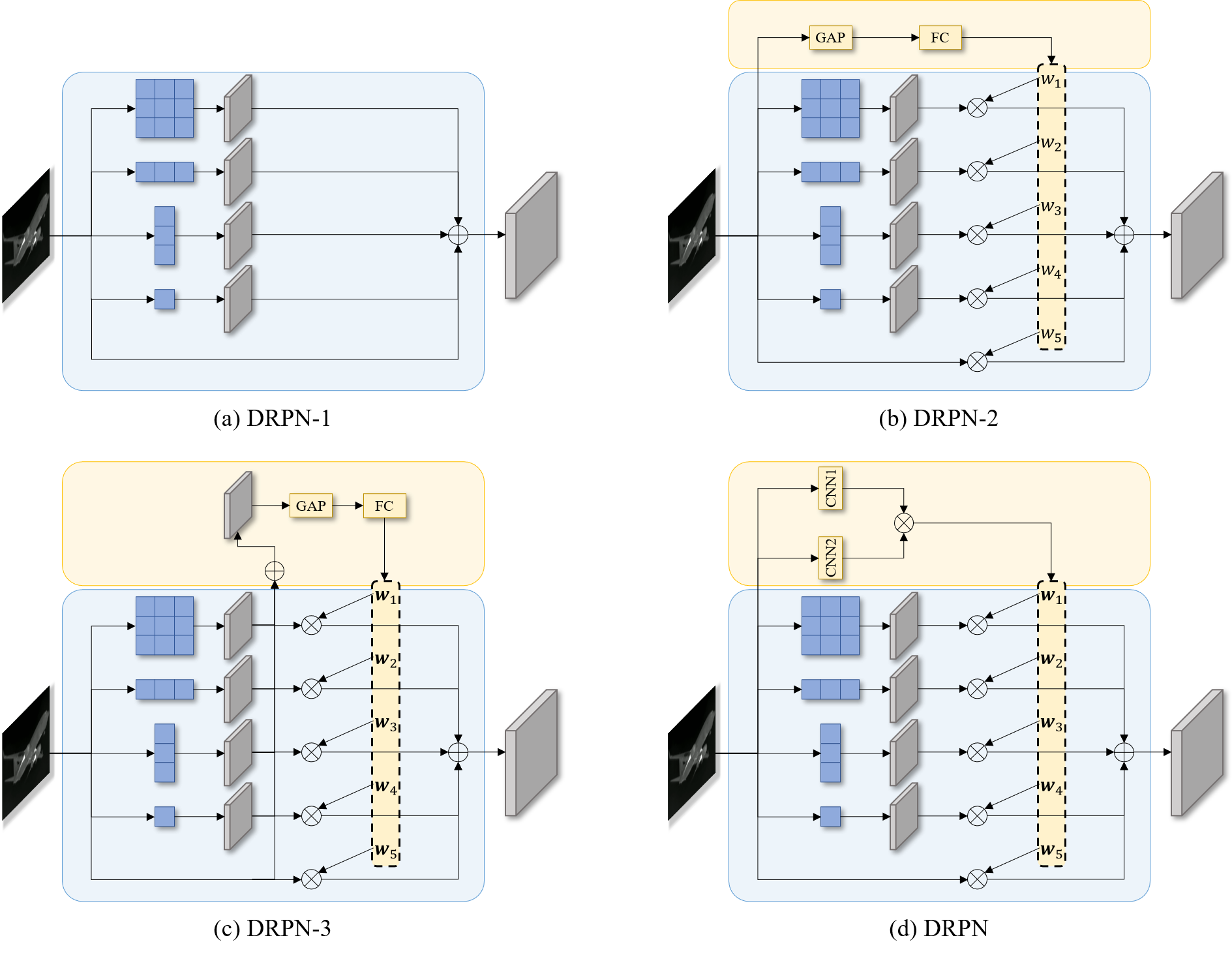}}
\par\end{centering}
\caption{\label{fig:4}Network structure diagram in ablation studies. The difference
is mainly that the weight generation modules are different. DRPN-1
has no weight generation module, DRPN-2 adopts the weight generation
module of dynamic convolution, and DRPN-3 adopts the weight generation
module of SKNet.}
\vspace{-0.5cm}\end{figure*}
\subsection{Ablation Studies\label{subsec:4.3}}
In order to validate the effectiveness of our major contributions,
we implement two variations, including 1) pruning multi-branch module;
2) pruning weight generation module, or changing the weight generation
module. The comparison results on InfraPlane dataset are shown in
Tab. \ref{tab:3}. Where SSD-vanilla represents the original SSD \cite{ssd}
detector without bells and whistles. DRPN-1 represents an SSD detector
with the multi-branch module, which is shown in Fig. \ref{fig:4}(a).
DRPN-2 represents SSD detector with multi-branch module and weight
generation module, and the weight generation module uses the corresponding
module in dynamic convolution \cite{dyconv}, which is shown in Fig. \ref{fig:4}(b).
DRPN-3 represents SSD detector with multi-branch module and weight
generation module, but the weight generation module uses the corresponding
module in SKNet \cite{sknet}, which is shown in Fig. \ref{fig:4}(c).
DRPN represents our proposed DRPN introduced in Sec. \ref{sec:3},
as shown in Fig. \ref{fig:4}(d).

The first, second, and fifth rows in Tab. \ref{tab:3} show that both
modules of DRPN can improve detection performance. Specifically, the
multi-branch module can increase 0.2 AP compared with SSD-vanilla.
Furthermore, the weight generation module can improve 1.7 AP on the
basis of the multi-branch module. The third, fourth, and fifth rows
of Tab. \ref{tab:3} show the detection results when using different
weight generation module methods. Our DRPN obtained the best detection
results.

In both DRPN-2 and DRPN-3, the three-dimensional tensor is compressed
into a one-dimensional vector by global average pooling, and then
the weights are generated by full connection layers. In DRPN-2, dynamic
convolution is adopted to compress the input features, while in DRPN-3,
SKNet is utilized to compress the output features. The other difference
is that the weights generated by dynamic convolution be $W^{DC}=[w_{1}^{DC},w_{2}^{DC},w_{3}^{DC},w_{4}^{DC},w_{5}^{DC}]^{T}\in\mathbb{R}^{5\times1}$,
i.e., each weight $w_{i}^{DC}\in\mathbb{R}(i\in{1,2,3,4,5})$ is a
scalar, while the weights generated by SKNet be $W^{SK}=[\boldsymbol{w}_{1}^{SK},\boldsymbol{w}_{2}^{SK},\boldsymbol{w}_{3}^{SK},\boldsymbol{w}_{4}^{SK},\boldsymbol{w}_{5}^{SK}]^{T}\in\mathbb{R}^{5\times C_{out}}$,
i.e., each weight $\boldsymbol{w}_{i}^{SK}\in\mathbb{R}^{C_{out}}(i\in{1,2,3,4,5})$
is a column vector whose dimension is equal to the number of output
channels $C_{out}$. These two modules are essentially for channel
weighting, which focus on the contribution of different channels.
The weight generation module of our DRPN is better than that of the two methods because the weight generation module of DRPN is essentially an uncompressed attention method, in which the feature map is not operated by global average pooling. Therefore the weight generation module of our DRPN focuses on the relationship between pixels, which has the ability to dynamically compute the weights for different convolutional kernels according to the target size.
\begin{table*}[!t]
\caption{\label{tab:3}Results on the infrared airplane dataset. Starting from
the baseline (SSD-vanilla), we gradually add the multi-branch module,
add and change the weight generation module in our SSD for ablation
studies.}

\centering{}\resizebox{\textwidth}{!}{
\begin{tabular}{ccc|cc|ccc}
\hline
Method  & Multi-branch Module  & Weight Generation Module  & AP  & AP50  & APs  & APm  & APl \tabularnewline
\hline
SSD-vanilla  & -  & -  & 37.4  & 79.6  & 33.6  & 63.5  & 81.2 \tabularnewline
DRPN-1  & \textsurd{}  & -  & 37.8  & 77.4  & 32.9  & 62.9  & 82.5 \tabularnewline
DRPN-2  & \textsurd{}  & Dynamic Conv  & 35.6  & 75.2  & 31.5  & 63.1  & 88.4 \tabularnewline
DRPN-3  & \textsurd{}  & SKNet  & 37.2  & 77.6  & 33.6  & 61.0  & 80.4 \tabularnewline
DRPN  & \textsurd{}  & uncompressed attention  & 39.5  & 81.5  & 35.8  & 65.1  & 83.3 \tabularnewline
\hline
\end{tabular}}
\vspace{-0.5cm}\end{table*}
\begin{table*}[!t]
\caption{\label{tab:4}The differences between weight generation methods.}

\centering{}\resizebox{\textwidth}{!}{
\begin{tabular}{ccccc}
\hline
Weight Generation Module  & Compression Method  & Weight Dimension  & Computation Order  & Number of Convolution \tabularnewline
\hline
\multirow{2}{*}{dynamic convolution \cite{dyconv}} & \multirow{2}{*}{global average pooling} & \multirow{2}{*}{$W^{DC}\in\mathbb{R}^{5\times1}$} & First generate weights,  & \multirow{2}{*}{1}\tabularnewline
 &  &  & then preform convolution.  & \tabularnewline
\multirow{2}{*}{SKNet \cite{sknet}} & \multirow{2}{*}{global average pooling} & \multirow{2}{*}{$W^{SK}\in\mathbb{R}^{5\times C_{out}}$} & First preform convolution,  & \multirow{2}{*}{5}\tabularnewline
 &  &  & then generate weights.  & \tabularnewline
\multirow{2}{*}{DRPN (Ours)} & \multirow{2}{*}{uncompressed attention} & \multirow{2}{*}{$W=\in\mathbb{R}^{5\times C_{out}}$} & First generate weights,  & \multirow{2}{*}{3}\tabularnewline
 &  &  & then preform convolution.  & \tabularnewline
\hline
\end{tabular}}
\vspace{-0.25cm}\end{table*}

In terms of computation order, SKNet first performs convolution, then
calculates the weight based on the output tensor, and finally weights
the output tensor, whose computational burden is large. Both dynamic
convolution and our DRPN first calculate the weight according to the
input tensor, then weight the convolution kernels, and finally perform
convolution. When in the inference phase, DRPN-2 adopted dynamic convolution
because the weight is calculated first, the multi-branch can be transformed
into a single-branch by re-parameterization technique, it only needs
one convolution calculation. DRPN-3 adopted SKNet because convolution
is performed first, the multi-branch cannot be transformed into a
single-branch, it needs five convolution calculations whose computational
burden is large. DRPN needs three convolution calculations (two from
attention, one from single-branch), which is a compromise
between DRPN-2 and DRPN-3, it can balance detection precision and
inference speed. In this way DRPN can achieve effective inference.
In summary, their differences are shown in Tab. \ref{tab:4}.

\begin{figure}[!t]
\begin{centering}
\textsf{\includegraphics[width=1\textwidth]{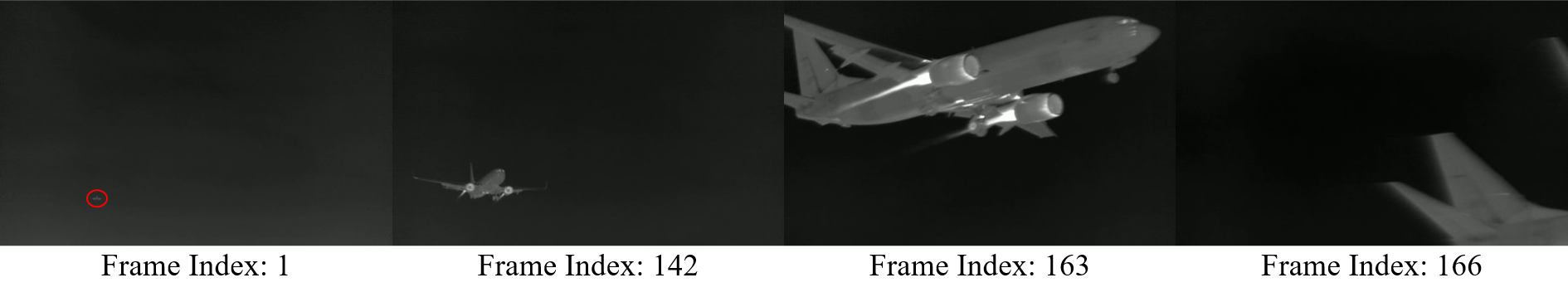}}
\par\end{centering}
\caption{\label{fig:5}Samples of the infrared image sequence in InfraPlane
datasets. In the sequence, the airplane is from far to near, and the
size of the target (airplane) is from small to large.}
\vspace{-5mm}\end{figure}
\subsection{Qualitative Performances\label{subsec:4.4}}
In order to show the effectiveness of DRPN in dynamically adjusting
the receptive field under the scale variation, we selected a sequence
of infrared images that contains an airplane flying from far to near,
then disappearing outside the image as shown in Fig. \ref{fig:5}.
The size of this airplane changes from small to large in this sequence.
The image size is 480\texttimes 300, the smallest target size is 5\texttimes 15,
and the largest target size is 461\texttimes 155.
\begin{figure}[thb]
\begin{centering}
\subfloat[The mean weights of 3\texttimes 3 convolution \\kernel.]{\begin{centering}
\includegraphics[width=0.5\textwidth]{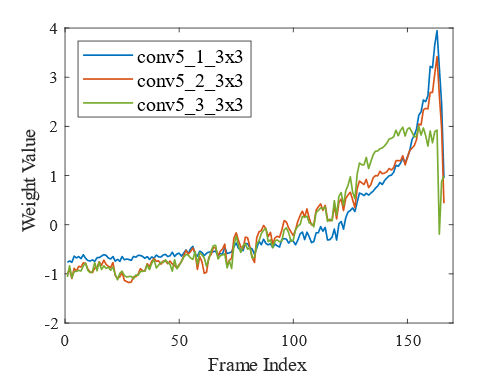}
\par\end{centering}
}\subfloat[The mean weights of 1\texttimes 1 convolution \\kernel.]{\centering{}\includegraphics[width=0.5\textwidth]{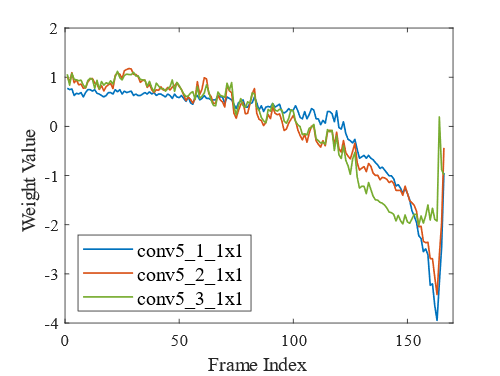}}
\par\end{centering}
\caption{\label{fig:6}The weight changes of convolution kernel branches. The
weights of the large convolution kernel increase with the increase
of target size; while the weights of the small convolution kernel
decrease with the increase of target size.}
\vspace{-0.5cm}\end{figure}

According to the experiments of Yang et al. \cite{condconv}, the distribution
of the weights is very similar across classes at early layers in the
network, and become more and more different at later layers. Therefore,
we select the weights of conv5\_1 \textasciitilde{} conv5\_3 in the
SSD. Fig. \ref{fig:6} to show the changing of the weights of different
branches in the inference of this sequence. It can be found from Fig.
\ref{fig:6}(a) that the weights of the large convolution kernels,
such as 3\texttimes 3 convolution kernel, become larger with the increase
of target size. The reason is that the 3\texttimes 3 convolution kernel
has a relatively bigger receptive field, which is more suitable for
detecting large targets. While in Fig. \ref{fig:6}(b) the weights
of the small convolution kernels decrease, that is because the small
convolution kernels, such as 1\texttimes 1 or shortcut branch, are
more suitable for small target detection. When the airplane finally
disappears out of the image, all the weights of all big convolution
kernels and small convolution kernels are close to zero. It means
that the output should approach zero if the target disappears.

Based on all the experiments performed in this section, we conclude
that:
\begin{enumerate}
\item The multi-branch module has different receptive fields for detecting
targets at different scales in infrared images.
\item The weight generation module can adaptively weight different branches
according to target scale changes. Both two modules can improve detection
precision.
\item Compared with the channel-level weight generation module adopted by
dynamic convolution and SKNet, the pixel-level weight generation module
utilized by DRPN can focus on the changing of the target size and
is more suitable for the serious scale variation problem in infrared
target detection.
\end{enumerate}

\section{Conclusions}
In this paper, we proposed a dynamic re-parameterization network (DRPN)
to deal with the problem of scale variation in infrared target detection.
DRPN consists of a weight generation module and a multi-branch module.
The multi-branch module has multiple branches with different receptive
fields. The weight generation module makes DRPN automatically weights
multiple branches. In virtue of multiple branches with different receptive
fields, the multi-branch module makes DRPN dynamically adjust its
receptive field. The multi-branch structure is used in the training
phase and then converted to a single-branch structure in the inference
phase to maintain effective inference. In the weight generation module,
dynamic convolution aggregates multiple convolution kernels to make
DRPN dynamically adjust its receptive field to adapt to each input
based on target sizes. We conducted extensive experiments on FLIR,
KAIST, and our InfraPlane datasets, and validated the effectiveness
of our method as well as the main modules therein. In the future,
we will try to combine DRPN with other scale variation methods to
further improve the detection performance.

\clearpage

\bibliographystyle{unsrt}
\bibliography{export}
\end{document}